# Image-Processing Based Methods to Improve the Robustness of Robotic Gripping

Kristóf Takács[1,2], Renáta Nagyné Elek[1,3], Tamás Haidegger[4,5]

*Abstract*—Image processing techniques have huge impact on most fields of robotics and industrial automation. Real time methods are usually employed in complex automation tasks, assisting with decision making or directly guiding robots and machinery, while post-processing is usually used for retrospective assessment of systems and processes. While artificial intelligence based image processing algorithms (usually neural networks) are more common nowadays, "classical" methods can also be used effectively even in modern applications. This paper focuses on optical flow based image processing, proving its efficiency by presenting optical flow based solutions for modern challenges in different fields of robotics such as robotic surgery and food industry automation. The main subject of the paper is a smart robotic gripper designed for automated robot cells in the meat industry, that is capable of slip detection and secure gripping of soft, slippery tissues with the help of the implemented optical flow based algorithm.

## I. Background

### A. Meat industry automation

Demand for automation in the red meat sector is steadily increasing due to growing consumer claims, which was also enhanced by the global COVID-19 pandemic [1]–[3]. Currently the advanced abattoirs are generally semi-automated plants, meaning that different machines – and sometimes robot arms – are working together with human operators to accomplish the different meat processing steps [4]. In semi-automated plants typically the simple, repetitive and/or force demanding tasks are handled by machines (e.g., cutting the carcass in half, placing final meat products in boxes), and the more complex steps are carried out by operators (e.g., separating meat from bones, cutting close to the intestines).

The difficulty of the full automation of a meat processing plant mainly derives from the natural diversity of animals, the extremely high risk of contamination (either from the intestines or from the skin surface) and the strict and inflexible regulation of the field [5]. The break-through towards completely automated solutions in the food industry started with intelligent robotic tools, that were first employed in the agriculture sector [6], [7]. These typically embedded systems are usually capable of sensing their environment using a number of different sensors, identifying and cutting/grasping/examining their target without any human intervention – or at most with remote supervision. The most regular sensors in these devices are different kinds of cameras (RGB-D, infrared, multi-spectral etc.) usually combined with an image processing deep neural networks (DNN), however classical image processing techniques – such as Hugh transformation or optical flow – can be used efficiently too.

To reduce barriers to automation, the RoBUTCHER concept was born, offering the so called meat factory cell (MFC) [8], [9]. Inside the MFC the primary slaughter steps are executed by two high-payload industrial robots with custom-developed smart end-of-arm-tooling (EOAT), supported by a motorized Carcass Handling Unit (CHU). The most important fields of research during the project are AI, cognitive systems and sustainability analysis (life cycle assessment) for this new slaughter concept [1]. The first two provide the necessary inputs for the robots to interact with the carcass during cutting, grasping and lifting tasks, employing concepts, methods and strategies from other domains, such as image-guided surgery [10].

Grasping and gripping are the most common tasks in industrial and service robotic applications, where physical interaction between the robot and its environment is required. Current challenges of robotic gripping include undefined target-object shapes and positions, delicate or highly elastic target materials, slippery or soft surfaces, and constantly changing or undefined environments [7]. These listed difficulties regularly occur in robotic surgery and agricultural robotics applications too, thus similar methods might be effective in both fields.

### B. Optical flow

Optical flow is a computer vision method, which is a pattern of motion of objects in a visual scene caused by the relative motion between the observer and the scene [11]. Optical flow is a fundamental algorithm for video analysis, especially in movement detection. It estimates motion between two consecutive video frames by calculating the pixel displacements [12]. Optical flow refers to the motion in the visual field based on pixel intensities. The fundamental assumption in optical flow

This work has received funding from the European Union's Horizon 2020 research and innovation programme under grant agreement No 871631, RoBUTCHER (A Robust, Flexible and Scalable Cognitive Robotics Platform). Kristof Takacs and Renáta Nagyné Elek acknowledges the financial support of Óbuda University - Doctoral School of Applied Informatics and Applied Mathematics.

[1] Antal Bejczy Center for Intelligent Robotics, Obuda Universtiy, Budapest, Hungary
[2] Doctoral School of Applied Informatics and Applied Mathematics, Óbuda University
[3] John von Neumann Faculty of Informatics, Obuda University, Budapest, Hungary
[4] University Research and Innovation Center (EKIK), Óbuda University, Budapest, Hungary
[5] Austrian Center for Medical Innovation and Technology, Wiener Neustadt, Austria

is that the intensity of the pixels is not changing during the motion—which is true in general, however it is not provided in the case of lighting changes.

The fundamental assumption of brightness constancy is the basis of optical flow:

$$I(x, y, t) = I(x + dx, y + dy, t + dt), \quad (1)$$

where x and y are the pixel coordinates, dx and dy are the pixel displacement, t is the time and dt is the elapsed time. From the Taylor expansion of this basic equation the optical flow constraint equation can be derived:

$$I_x v_x + I_y v_y + I_t = 0, \quad (2)$$

where $I_x$, $I_y$ and $I_t$ are the derivatives of the image function $I(x, y)$ with respect to x, y and t; vector $V = (v_x, v_y)$ defines the velocity vector in x and y direction. This equation is called the optical flow equation.

Nevertheless, beyond the brightness constancy assumption, the optical flow equation is still under-determined; thus, the different approaches use further restrictions, e.g., that the intensity of the local neighbors of pixels changes similarly.

There are two main categories of optical flow techniques: dense and sparse methods. Dense optical flow algorithms calculate optical flow for all pixels, sparse techniques calculate the flow just for some pixels (special features, such as corners, edges). The Lucas–Kanade method is commonly used to calculate optical flow for only special features. The main idea of this method is based on a local motion constancy assumption, where nearby pixels have the same displacement direction. Based on the different approaches, dense optical flow is more accurate, but naturally, it needs more computational capacity [13].

For this work, the chosen technique was a dense optical flow method, the Farneback optical flow. The Farneback method has high accuracy, and it is useful to examine all of the pixels in the image. Farneback algorithm is a two-frame optical flow calculation technique that uses polynomial expansion, where a polynomial approximates the neighborhood of the image pixels. Quadratic polynomials give the local signal model represented in a local coordinate system [14].

## II. OPTICAL FLOW IN MODERN ROBOTICS

### A. Robotic surgery applications

The optical flow method is widely used in different applications where image processing is useful or required. One of its most promising utilization is in the field of surgical robotics, where the digital video feed of the endoscope (or stereo-endoscope) is always accessible. Although the video of a surgical procedure can be processed in real-time too, the vast majority of publications use recorded videos or previously published datasets (e.g., the JIGSAWS - JHU-ISI Gesture and Skill Assessment Working Set [15]).

Employing robots, machines and modern digital technologies in the surgical rooms had huge impact not only on the outcome of the surgeries, but also on surgical skill assessment and analysis of the surgeons' work [16], [17]. Although artificial intelligence based methods (e.g., Deep Neural Networks) are the most common in automated video processing, classic image processing approaches and optical flow based methods can be used with good results too.

Sarikaya et al. in [18] propose and optical flow based method for surgical gesture recognition which is usually done by analysing the kinematic data of the surgical robot. The described method relies exclusively on the observable motion on surgical videos i.e., dense optical flow data and shows that this can be a robust alternative to kinematic data analysis. The accuracy of their Optical flow ConvNet system on the JIGSAWS dataset was between 74% and 92% for regular surgical gestures.

Liu et al. in [19] present a real-time utilization of optical flow calculation of a surgical video-feed. In their work the authors prove the effectiveness of an advanced optical flow based algorithm for tracking colonoscopy procedures by compering real-life videos with videos from virtual and modeled environments. The paper shows the robustness of the technique against fluid and illumination artifacts, blurry images and structural changes too. Furthermore, the merged use of sparse and dense optical flow fields also improves the performance by allowing to compute the focus of expansion.

Other researches show that real-time optical flow calculation can be a powerful tool in therapy too. Zachiu et al. proposed an optical flow based tracking solution to improve MR-guided beam therapies, described in [20]. Their work focuses on solving one of the biggest challenges of any beam therapies, the tracking of organ-movements during the procedure. Zachiu et al. use optical flow to determine and separate movements within the target region and in the vicinity of the target region in real time e.g., movements originating from the pulsing of different arteries within the target organ or close to it. Their experiments show that improving and optimizing the existing optical flow methods for given tasks can be highly beneficial.

In the case of Minimally Invasive Surgery (MIS) and Robot-Assisted Minimally Invasive Surgery (RAMIS) the surgeon reaches the inside organs through small skin incisions, and the operating area is visualized by an endoscopic camera.

The endoscopic camera provides mainly two dimensional image streams for MIS and three dimensional for RAMIS. Endoscopic images are the only sensory data in the case of MIS and an important sensory data in the case of RAMIS, besides kinematic data. Since images are available and easily recordable for both MIS and RAMIS, computer vision takes a huge role in surgical data science and automation. Optical flow usage in MIS and RAMIS can be found in several studies, such as in [21], where semantic surgical tool segmentation in endoscopic images was proposed, which can be an important step towards pose estimation, task automation and skill assessment in MIS operations. Here an efficient ground truth labelling method was proposed with the help of the optical flow algorithm for the JIGSAWS dataset [22], which is the most studied dataset for surgical skill assessment. The ground truth dataset and source codes are publicly available on Github

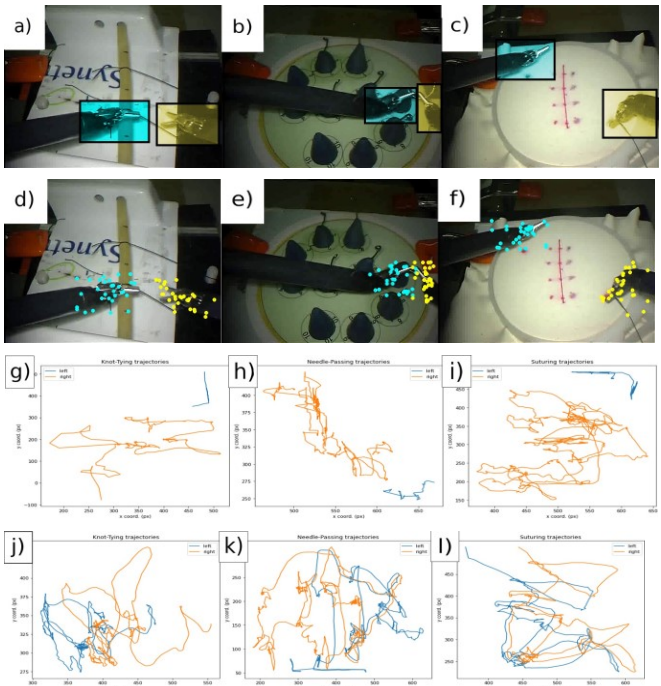

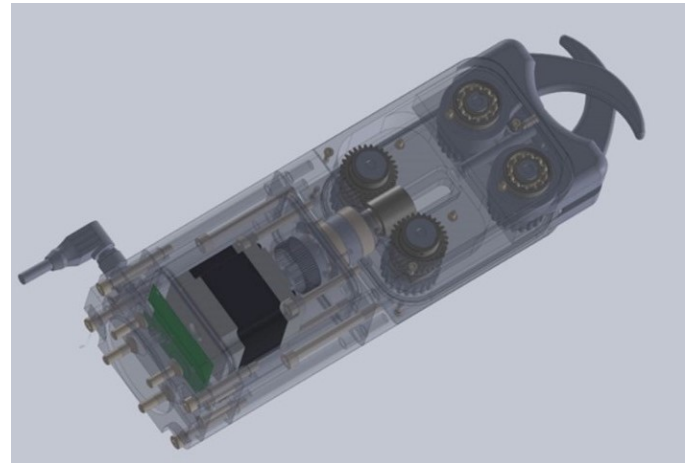

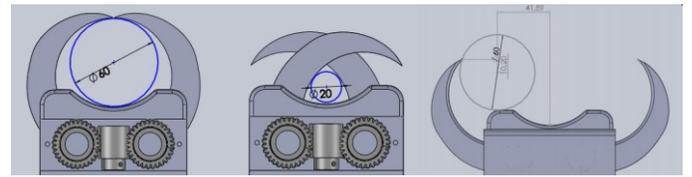

Fig. 1: Surgical tasks in JIGSAWS: Knot-Tying, Needle-Passing and Suturing. (**a-c**) Region of Interest manual selection, (**d-f**) The initial samples computed by the optical flow algorithm. The third an fourth row show one tracked points' trajectory from each surgical tool (blue=left and orange=right). (**g-i**) the data is from a novice user (**j-l**) it is an expert subject.

Fig. 2: (a) Shows the CAD model of the smart gripper used in the RoBUTCHER project. (b) Shows its closing motion as an example of "robustness by design", the encircling–centering motion tolerates high variance of target shape, size and displacement [23].

(https://github.com/dorapapp96/SurgToolSegJIGSAWS.git). Objective skill assessment based personal performance feedback is a crucial part of surgical training, however it is not the part of the clinical practice yet. In [13], the potential usage of optical flow as an input for RAMIS skill assessment was shown, including the maximum accuracy achievable with this image-based data by evaluating it with an established skill assessment benchmark, by evaluating its methods independently, what outperformed the state of the art (Fig. 1).

*B. Optical flow based slip detection*

Within the field of food industry, the automation of meat processing plants also has the potential to benefit from optical flow based methods. The most significant challenges in meat-industry automation – similarly to most fields of the food and agriculture industry – is the handling of high variance coming from natural biodiversity, and the detection and management of diseased and abnormal animals.

Any machinery and robot system designed for meat industry automation must be able to autonomously deal with this outstandingly high variance of mechanical characteristics (size, weight, elasticity, surface properties etc.) of the "target objects" (animal carcasses, parts of animals, pieces of meat, organs etc.). One solution is to make the devices tolerant to high variability "by design". This means, that the mechanical design of the device is capable of fulfilling its tasks without any sensors or "intelligence" regardless of the uncertain and/or unknown characteristics and coordinates of the target object (until they stay between reasonable limits). Fig. 2b shows an example meat industry gripper that is capable of gripping target objects between 2–6 cm up to a displacement of 4 cm thanks to the encircling motion of the gripping fingers.

The more advanced solution for improving robustness is the use of sensors and built-in intelligence. In this case, the most important feature of the tools is not the hardware design, but the intelligent components and the software behind them. Regarding the place of the decision making there are two main options in food industry automation [24]. Huge group of robots working in bounded environments usually make a centralized system, where a central compute module makes the high-level decisions, communicates with and controls all robots and devices. Creating so called embedded systems is the other solution, where the robots or devices process all the sensory data internally and make all the low and high level calculations and decisions by itself.

In the RoBUTCHER project – described in Section I-A – these two methods are mixed. The robot system is fundamentally centralized, a central computer collects most of the sensor data, runs the Virtual Reality module, AI modules and controls the robots. On the other hand, the smart end of arm tools (gripper and knife) work as embedded systems with own dedicated communication lines and/or integrated compute modules. One of the robotic grippers in the RoBUTCHER project is a smart gripper with integrated microcontroller, sensors and a Raspberry Pi Zero, the mechanical design is shown on Fig. 2a [23]. Its mechanism provides encircling gripping motion, fail-safe grasping and high clamping force by design, however the integrated sensors and built-in intelligence are the features that make the gripper unique and capable of accomplishing the complex tasks required for entirely automated meat-processing.

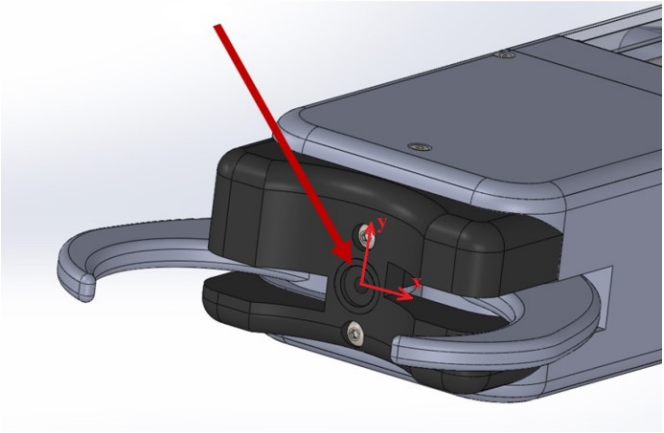

Fig. 3: Frontal view of the gripper in an open state. The endoscopic camera is placed inside the black "passive finger", the synchronized encircling motion of the active fingers push the target object guide and press the target objects to the lens in front of the camera. The lens is needed to keep the gripper waterproof and to maintain the 2 cm focal length between the camera and the surface of the target object.

One of the main smart features of the gripper is the optical flow based slip detection. Detecting slippage is a crucial feature of any automation development where gripping is required. This is especially true in the field of food industry, since the target objects are usually food products that suffer contamination in case of slipping and can't be sold anymore. In addition – compared to other fields of industry – slippage is rather frequent during food industry automation because of the slippery, wet, soft surfaces and undefined, varying target shapes and dimensions.

The hardware side of the slippage detection system in the smart gripper consists of an endoscopic camera (640x480 pixel, 30 fps, 2 cm focus, 70° view angle, built-in leds) and a Raspberry Pi Zero W. The camera is placed at the front of the gripper behind a protective lens as shown on Fig. 3. Because of the synchronized encircling motion of the fingers, the target object is always centered when a stable grasp is reached, thus it is tightly pressed against the protective lens in front of the camera. This design ensures that the surface of the target object is exactly in the focal point of the endoscopic camera, providing a clear view of the target. The illumination is provided by the built-in leds of the endoscopic camera.

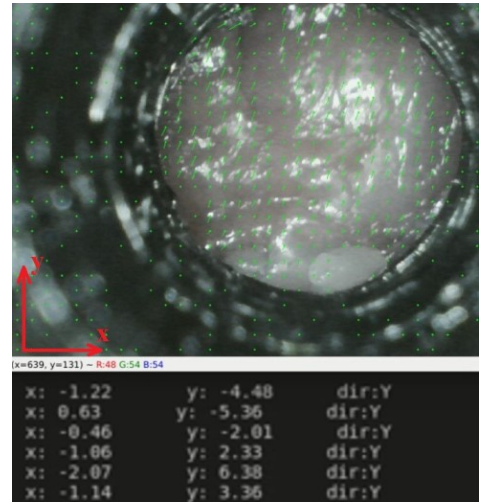

(a)

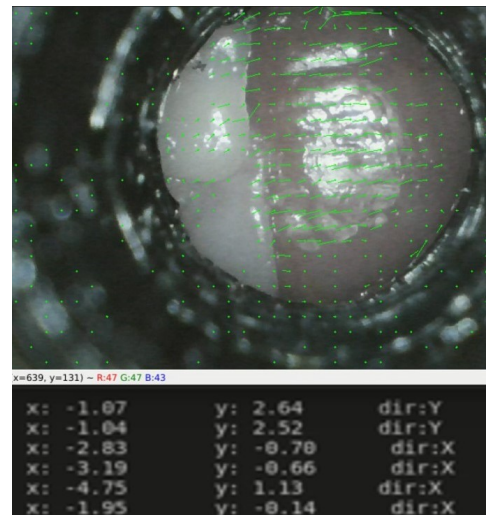

(b)

Fig. 4: Visualization of the optical flow on the images of the built-in camera with green vectors. The black housing of the camera is static on the video feed (hence the green dots), while the grasped meat product is moving in different directions. (a) shows real slippage i.e., target movement in the Y direction, while (b) shows "rotating" movement of grasped target which – in this case – requires no intervention.

The optical flow software is written in Pyhton depending on OpenCV, using the Farneback optical flow algorithm. Fig. 4 shows typical frames from the endoscopic camera, however they are already cropped and their resolution is reduced for runtime optimization. The Farneback algorithm produces a 3D optical flow matrix with the calculated direction of movement of the pixels (visualised by green vectors on Fig. 4). The image can be divided into two main regions, the ROI (Region of Interest) is the circle in which the target object can be seen, while the region outside of the circle is the housing of the camera, which is never moving, thus the optical flow of those pixels are – hypothetically – always zero.

Although the target objects of the gripper are soft and easily deformable pieces of tissue, it is hypothesized, that the small piece of surface in the RoI (about 2 cm in diameter) that the camera can see does not suffer significant deformation, thus the camera sees a rigid body. This assumption means, that the points inside the RoI always have the same speed, thus for further analysis the average of the optical flow field is being used, representing the actual velocity of the target object.

As described in [23] the gripper was designed to be used for various gripping tasks within automated pig slaughtering, which is supported by variable clamping force (i.e., force controlled grasping). In the ideal case the target object is grasped tightly and remains in a relative rest throughout the whole process, thus the estimated velocity remains zero. However, when the object is slipping the 'y' component of the estimated velocity becomes higher (see Fig 4a). In this case, the gripper can either react internally by increasing its own clamping force, or can simply send an error message to the robot cell (and/or to the supervising operator) that the target might have slipped out and the process should be stopped. If movement along the 'x' axis is detected, it only means that the object is rotating within the grasp of the gripper and no intervention is required.

### III. Discussion

This paper presented various scenarios where classical image processing techniques – such as optical flow calculation – can be used effectively. Although AI based image processing became rather popular in the recent years, hard computing methods are still viable and sometimes even more effective. Optical flow is a good example for such methods, several different algorithms are implemented in the most commonly used image processing libraries.

In this paper several examples were given from modern fields of robotics where optical flow based algorithms has great potential. In Robotic surgery optical flow is usually used on recorded videos and datasets for surgical skill assessment, however real-time applications for tracking purposes were also identified. In contrast, the food industry applications of optical flow algorithms are usually real-time, since it is a powerful tool for fast detection of slippage, which is a frequent problem regarding automated grasping.

### References

[1] C. Valente, H. Møller, F. M. Johnsen, S. Saxegårrd, E. R. Brunsdon, and O. A. Alvseike, "Life cycle sustainability assessment of a novel slaughter concept," *Journal of Cleaner Production*, vol. 272, p. 122651, 2020.

[2] B. Blagojevic, T. Nesbakken, O. Alvseike, I. Vågsholm, D. Antic, S. Johler, K. Houf, D. Meemken, I. Nastasijevic, M. V. Pinto *et al.*, "Drivers, opportunities, and challenges of the european risk-based meat safety assurance system," *Food Control*, p. 107870.

[3] A. Khamis, J. Meng, J. Wang, A. T. Azar, E. Prestes, Á. Takács, I. J. Rudas, and T. Haidegger, "Robotics and intelligent systems against a pandemic," *Acta Polytechnica Hungarica*, vol. 18, no. 5, pp. 13–35, 2021.

[4] G. Purnell and G. I. of Further, "Robotics and automation in meat processing," in *Robotics and Automation in the Food Industry*. Elsevier, 2013, pp. 304–328.

[5] K. Takács, A. Mason, L. E. Cordova-Lopez, and T. Haidegger, "Open issues in agri-food robot standardization—the red meat sector," in *2021 IEEE 15th International Symposium on Applied Computational Intelligence and Informatics (SACI)*. IEEE, 2021, pp. 000 363–000 368.

[6] B. Zhang, Y. Xie, J. Zhou, K. Wang, and Z. Zhang, "State-of-the-art robotic grippers, grasping and control strategies, as well as their applications in agricultural robots: A review," *Computers and Electronics in Agriculture*, vol. 177, p. 105694, 2020.

[7] K. Takács, A. Mason, L. B. Christensen, and T. Haidegger, "Robotic grippers for large and soft object manipulation," in *2020 IEEE 20th International Symposium on Computational Intelligence and Informatics (CINTI)*. IEEE, 2020, pp. 133–138.

[8] O. Alvseike, M. Prieto, K. Torkveen, C. Ruud, and T. Nesbakken, "Meat inspection and hygiene in a meat factory cell – an alternative concept," *Food Control*, vol. 90, pp. 32–39, 2018.

[9] I. de Medeiros Esper, A. Mason *et al.*, "Robotisation and intelligent systems in abattoirs," *Trends in Food Science & Technology*, vol. 111, no. XX, pp. 1–15, 2021.

[10] T. Haidegger, "Autonomy for Surgical Robots: Concepts and Paradigms," *IEEE Transactions on Medical Robotics and Bionics*, vol. 1, no. 2, pp. 65–76, May 2019.

[11] D. Sun, S. Roth, and M. J. Black, "Secrets of optical flow estimation and their principles," in *2010 IEEE Computer Society Conference on Computer Vision and Pattern Recognition*, Jun. 2010, pp. 2432–2439.

[12] A. I. Karoly, R. N. Elek, T. Haidegger, K. Szell, and P. Galambos, "Optical flow-based segmentation of moving objects for mobile robot navigation using pre-trained deep learning models," in *2019 IEEE International Conference on Systems, Man and Cybernetics (SMC)*. Bari, Italy: IEEE, 10 2019, pp. 3080–3086.

[13] G. Lajkó, R. Nagyné Elek, and T. Haidegger, "Endoscopic image-based skill assessment in robot-assisted minimally invasive surgery," *Sensors*, vol. 21, no. 16, p. 5412, 2021.

[14] G. Farnebäck, "Two-Frame Motion Estimation Based on Polynomial Expansion," in *Image Analysis*, G. Goos, J. Hartmanis, J. van Leeuwen, J. Bigun, and T. Gustavsson, Eds. Berlin, Heidelberg: Springer Berlin Heidelberg, 2003, vol. 2749, pp. 363–370.

[15] Y. Gao, S. S. Vedula, C. E. Reiley, N. Ahmidi, B. Varadarajan, H. C. Lin, L. Tao, L. Zappella, B. Béjar, D. D. Yuh *et al.*, "Jhu-isi gesture and skill assessment working set (jigsaws): A surgical activity dataset for human motion modeling," in *MICCAI workshop: M2cai*, vol. 3, no. 3, 2014.

[16] T. Haidegger, S. Speidel, D. Stoyanov, and R. M. Satava, "Robot-assisted minimally invasive surgery—surgical robotics in the data age," *Proceedings of the IEEE*, vol. 110, no. 7, pp. 835–846, 2022.

[17] R. Elek and T. Haidegger, "Robot-Assisted Minimally Invasive Surgical Skill Assessment—Manual and Automated Platforms," *Acta Polytechnica Hungarica*, vol. 16, no. 8, 9 2019.

[18] D. Sarikaya and P. Jannin, "Surgical gesture recognition with optical flow only," *arXiv preprint arXiv:1904.01143*, 2019.

[19] J. Liu, K. R. Subramanian, and T. S. Yoo, "An optical flow approach to tracking colonoscopy video," *Computerized Medical Imaging and Graphics*, vol. 37, no. 3, pp. 207–223, 2013.

[20] C. Zachiu, N. Papadakis, M. Ries, C. Moonen, and B. D. De Senneville, "An improved optical flow tracking technique for real-time mr-guided beam therapies in moving organs," *Physics in Medicine & Biology*, vol. 60, no. 23, p. 9003, 2015.

[21] D. Papp, R. Nagyné Elek, and T. Haidegger, "Surgical tool segmentation on the jigsaws dataset for autonomous image-based skill assessment," in *IEEE 10th Jubilee International Conference on Computational Cybernetics and Cyber-Medical Systems ICCC 2022*. IEEE, 2022, p. 7.

[22] A. K. Lefor, K. Harada, A. Dosis, and M. Mitsuishi, "Motion analysis of the JHU-ISI Gesture and Skill Assessment Working Set using Robotics Video and Motion Assessment Software," *International Journal of Computer Assisted Radiology and Surgery*, vol. 15, no. 12, pp. 2017–2025, 12 2020. [Online]. Available: http://link.springer.com/10.1007/s11548-020-02259-z

[23] B. Takács, K. Takács, T. Garamvölgyi, and T. Haidegger, "Inner organ manipulation during automated pig slaughtering—smart gripping approaches," in *2021 IEEE 21st International Symposium on Computational Intelligence and Informatics (CINTI)*. IEEE, 2021, pp. 000 097–000 102.

[24] J. Billingsley, A. Visala, and M. Dunn, "Robotics in agriculture and forestry," 2008.